%% file: acl2023.tex
\documentclass[11pt]{article}
\usepackage{ACL2023} 
\usepackage{times}
\usepackage{blindtext}
\usepackage{latexsym}
\usepackage{multirow}
\usepackage{multicol}
\usepackage{amsmath}
\usepackage{booktabs}
\usepackage{xcolor}
\usepackage{float}
\usepackage{algpseudocode}
\usepackage{listings}
\usepackage{color}
\usepackage{amssymb}
\usepackage{footnote}
\usepackage{supertabular}
\usepackage{titlesec}
\usepackage{makecell}
\usepackage[toc,page,header]{appendix}

\DeclareCaptionLabelFormat{andtable}{#1~#2  \&  \tablename~\thetable}
\usepackage{graphicx}
\usepackage[flushleft]{threeparttable}
\definecolor{backcolour}{rgb}{0.95, 0.95, 0.96}
\lstdefinestyle{mystyle}{
    backgroundcolor=\color{backcolour},   
    showtabs=false,                  
    tabsize=2
}
\usepackage{enumerate}
\usepackage{enumitem}
\newlist{compactitem}{itemize}{3}
\setlist[compactitem]{topsep=0pt,partopsep=0pt,itemsep=0pt,parsep=0pt,leftmargin=\parindent}
\setlist[compactitem,1]{label=\textbullet}
\setlist[compactitem,2]{label=---}
\setlist[compactitem,3]{label=*}

\newlist{compactdesc}{description}{3}
\setlist[compactdesc]{topsep=0pt,partopsep=0pt,itemsep=0pt,parsep=0pt}

\newlist{compactenum}{enumerate}{3}
\setlist[compactenum]{topsep=0pt,partopsep=0pt,itemsep=0pt,parsep=0pt}
\setlist[compactenum,1]{label=\arabic*.}
\setlist[compactenum,2]{label=\alph*.}
\setlist[compactenum,3]{label=\roman*.}

\usepackage{longtable}
\usepackage[T1]{fontenc}
\usepackage[utf8]{inputenc}
\usepackage{microtype}
\usepackage{inconsolata}
\usepackage[capitalize,nameinlink]{cleveref}
\usepackage{subcaption}
\makeatletter
\renewcommand{\sectionautorefname}{\S\@gobble}
\renewcommand{\sectionautorefname}{\S\@gobble}
\renewcommand{\subsectionautorefname}{\S\@gobble}
\renewcommand{\sectionautorefname}{\S\@gobble}
\renewcommand{\subsectionautorefname}{\S\@gobble}
\makeatother

\usepackage{graphicx}
\usepackage{tikz}
\usepackage{forest}
\usepackage{tikz-qtree}
\usetikzlibrary{trees,positioning,shapes,shadows,arrows.meta}

\newcommand{\macro}[1]{\textcolor{black}{#1}} 

\newcommand{\task}{\macro{DIALECT-COPA}}
\newcommand{\gmnlp}{\macro{\textsc{GmuNLP}}}
\title{Data-Augmentation-Based Dialectal Adaptation for LLMs}

\author{Fahim Faisal, Antonios Anastasopoulos\\
Department of Computer Science, George Mason University\\
\texttt{\{ffaisal,antonis\}@gmu.edu}}

\begin{document}
\maketitle
\begin{abstract}
This report presents \gmnlp's participation to the \task{} shared task at VarDial 2024~\cite{chifu-etal-2024-vardial}, which focuses on evaluating the commonsense reasoning capabilities of large language models (LLMs) on South Slavic micro-dialects. The task aims to assess how well LLMs can handle non-standard dialectal varieties, as their performance on standard languages is already well-established. We propose an approach that combines the strengths of different types of language models and leverages data augmentation techniques to improve task performance on three South Slavic dialects: Chakavian, Cherkano, and Torlak. We conduct experiments using a language-family-focused encoder-based model (BERTić) and a domain-agnostic multilingual model (AYA-101). Our results demonstrate that the proposed data augmentation techniques lead to substantial performance gains across all three test datasets in the open-source model category. This work highlights the practical utility of data augmentation and the potential of LLMs in handling non-standard dialectal varieties, contributing to the broader goal of advancing natural language understanding in low-resource and dialectal settings.\footnote{Code and data are publicly available: \url{https://github.com/ffaisal93/dialect_copa}}

\end{abstract}

\section{Introduction}

Recent advancements in large language models (LLMs) have led to remarkable performance on a wide range of natural language understanding tasks, particularly in standard languages. However, the effectiveness of these models on non-standard dialectal varieties remains an open question~\cite{faisal2024dialectbench}. The \task{} shared task, introduced by \citet{ljubesic-etal-2024-dialect}, aims to bridge this gap by evaluating the commonsense reasoning capabilities of LLMs on South Slavic dialects.

Commonsense reasoning, as originally proposed by \citet{Gordon2011ChoiceOP}, requires models to make plausible inferences based on everyday knowledge and understanding of the world. Extending this task to dialects poses unique challenges, as models must capture the nuances and variations specific to these language varieties. The \task{} shared task provides a platform to explore the adaptability and generalization capabilities of LLMs in this context. 

In this \gmnlp{} submission, we explore the potential of data augmentation techniques in enhancing the performance of language models on dialectal commonsense reasoning tasks. Our approach harnesses the power of state-of-the-art LLMs to generate synthetic training data, which we combine with the provided training dataset. By employing a diverse set of language models, we aim to quantify the performance gains achievable through data augmentation. Specifically, we utilize three categories of language models to maximize dialectal task performance: (1) smaller language models that are well-suited for low-resource settings and can be easily customized, (2) mid-size language models that strike a balance between task-specific performance and language understanding capabilities, and (3) closed-source language models that generate high-quality synthetic task data to further enhance the performance of the other two categories of language models.

We achieved the highest scores across all three test datasets in the open-source model category. In addition, our solution performed on par with the GPT-4 zero-shot iterative prompting approach employed by one of the teams, demonstrating the competitiveness of the proposed approach against state-of-the-art closed-source models. Furthermore, we achieved substantial performance improvements for the small-scale, language-family-focused model BERTić by combining it with our data augmentation strategy, showcasing the effectiveness of our approach in boosting the performance of language models tailored for low-resource settings.

The remainder of this paper is organized as follows: \cref{sec:task} provides an overview of the \task{} shared task and dataset, \cref{sec:exp_phase} describes our methodology and experimental setup, \cref{sec:result} presents our results and analysis, and \cref{sec:conclusion} concludes the paper and discusses future directions.

\section{The DIALECT-COPA shared task}
\label{sec:task}
\paragraph{Task Information} In the \task{} shared task, a premise sentence is provided along with a question that can be either a cause or an effect. The objective is to build a classifier that selects the most plausible response from two candidate answer choices based on the given premise and question. To illustrate, consider the following training example in English, where the task is to identify the most plausible cause:
\\

\noindent\fbox{
\begin{minipage}{\columnwidth}
\small
\{"premise": "My body cast a shadow over the grass.", 

"choice1": "The sun was rising.", 

"choice2": "The grass was cut.", 

"question": "cause", "label": 0, "idx": 0\}
\end{minipage}}
\\
\\
The \task{} dataset consists of such cause-effect examples across 8 languages and dialects, challenging models to perform commonsense reasoning in non-standard language varieties.

\begin{table}[h]
\centering
\small
\begin{tabular}{p{1.4cm}p{3.5cm}@{}rrr@{}}
\toprule
code & language & train & val. & test \\
\midrule
en & English & 400 & 100 & \\
sl & Slovenian & 400 & 100 & \\
sl-cer & Cerkno & 400 & 100 & 500 \\
hr & Croatian & 400 & 100 & \\
hr-ckm & Chakavian & - & - & 500 \\
sr & Serbian & 400 & 100 & \\
sr-trans & Serbian (transliterated) & 400 & 100 & \\
sr-tor & Torlak & 400 & 100 & \\
sr-tor-trans & Torlak (transliterated) & 400 & 100 & 500 \\
mk & Macedonian & 400 & 100 & \\
mk-trans & Macedonian (transliterated) & 400 & 100 & \\
\bottomrule
\end{tabular}
\caption{\task{} dataset statistics for different languages and their dialectal varieties.}
\label{tab:dataset_stats}
\end{table}

\paragraph{Languages} The \task{} dataset encompasses training and validation data in 7 languages, including English, 6 moderately resourced South Slavic languages, and two related micro-dialects. The test dataset features these two micro-dialects along with an additional previously unseen dialect. The three dialects in the test set are as follows:

\begin{enumerate}
\item The Cerkno dialect of Slovenian, spoken in the Slovenian Littoral region, specifically in the town of Idrija.
\item The Chakavian dialect of Croatian from the northern Adriatic, particularly from the town of Žminj.
\item The Torlak dialect, spoken in southeastern Serbia, northeastern North Macedonia, and northwestern Bulgaria, with the specific test instances coming from the town of Lebane.
\end{enumerate}

Cerkno and Torlak dialects are present in all three dataset splits (training, validation, and test) whereas, the Chakavian dialect is intentionally held out from the training and validation splits and is exclusively encountered during the test phase. Each dialect in the test dataset comprises 500 instances. \cref{tab:dataset_stats} presents the detailed statistics of the \task{} dataset, providing an overview of the distribution of instances across languages and dialects.

\section{Experimental Phases}
\label{sec:exp_phase}
In this section, we report different phases of our experiments. We step by step perform experiments to choose appropriate base models followed by data augmentation, combination and task-specific model tuning.
\begin{table}
\centering
\tiny
\begin{tabular}{lp{2.5cm}lll}
\toprule
\textbf{Base model} & \textbf{Fine-tuning (FT)/Prompting} & \textbf{Epoch} &\multicolumn{2}{c}{\makecell{\textbf{Acc. (\%)}}} \\
\cmidrule{4-5}
&  & & \textbf{en} & \textbf{hr} \\
\midrule
Aya-101 & 4 shot (2 cause, 2 effect) &-& \textbf{80} & \textbf{75} \\
MaLA-500 & 4 shot (2 cause, 2 effect) &-& 50 & -- \\
Llama2-CHAT (7B) & 4 shot (2 cause, 2 effect) &-& 75 & 50 \\
\midrule
BERT & FT (eval. lang)  &3& \textbf{66} & 55 \\
mBERT & FT (eval. lang) &3& 55 & 57 \\
XLM-R & FT (eval. lang) &3& 54 & 54 \\
{BERT}i{\'c} & FT (eval. lang) &3& 48 & \textbf{64} \\
\bottomrule
\end{tabular}
\caption{Preliminary evaluation results on the English and Croatian validation set for different base models.}
\label{tab:preliminary_experiments}
\end{table}

\begin{table*}[htbp]
\centering
\small
\begin{tabular}{lp{8cm}l}
\toprule
\textbf{Data Identifier} & \textbf{Description} & \textbf{Covered Language} \\
\midrule
\texttt{[lang]-train} &  Original \task{} training data & en, hr, mk, sl, sl-cer, sr, sr-tor \\
\texttt{[lang]-trans} & Transliterated (Cyrillic $\rightarrow$ Latin) training data & mk, sr, sr-tor \\
\texttt{[lang]-claude}  & Providing grammar rules and few-shot Croatian-Chakavian examples to generate synthetic parallel hr-ckm-train examples given the hr-train examples  & hr-ckm \\
\texttt{[lang]-gpt4} &  Additional synthetic English training data generated by GPT-4~\cite{whitehouse-etal-2023-llm} & en \\
\texttt{[lang]-reverse} & Reverse-augmentation on \texttt{[lang]-train}, \texttt{[lang]-trans}, and \texttt{[lang]-claude} data & en, hr, mk, sl, sl-cer, sr, sr-tor \\
\texttt{[lang]-nllb}  & Machine translation of \texttt{en-gpt4} source data to other languages using the NLLB-6B model & hr, mk, sl,  sr\\
\bottomrule
\end{tabular}
\caption{Training data augmentation approaches}
\label{tab:data_augmentation}
\end{table*}

\paragraph{Phase 1: Model Selection} In the preliminary phase of our experiments, we conduct a series of trials to identify base language models that demonstrate strong performance on language understanding tasks in a multilingual context. To achieve this, we fine-tune widely-used encoder-based models, such as BERT \citep{devlin-etal-2019-bert}, mBERT, and XLM-R \citep{conneau-etal-2020-unsupervised}, on the English and Croatian subsets of the \task{} training dataset. Additionally, we explore the potential of more recently open-sourced large language models (LLMs) of varying sizes, such as LLaMA-2 \citep{touvron2023llama}, Aya-101~\cite{ustun2024aya} and MaLA-500 \citep{lin2024mala500}, to gauge their effectiveness on the task.

Our key observations from this preliminary phase are as follows:
\begin{enumerate}[label=\textrightarrow, leftmargin=*]
    \item BERT, mBERT, and XLM-R exhibit comparable performance on the Croatian subset, achieving an accuracy of around 55\%(+/-) after 3 epochs of in-language fine-tuning. However, the monolingual English BERT model surpasses the multilingual models on the English subset when fine-tuned for the same number of epochs.
    \item {BERT}i{\'c} \citep{ljubesic-lauc-2021-bertic}, a transformer-based model pre-trained on Bosnian, Croatian, Montenegrin, and Serbian languages, aligns well with the target languages of the \task{} test set. Fine-tuning BERTi\'c on the Croatian subset yields a notable performance improvement of approximately 12 percent (i.e. 7 percentage points) compared to the aforementioned multilingual models.
    \item Employing 4-shot prompting with the LLaMA-2 7B parameter model results in better performance on the English subset. However, for the Croatian subset, LLaMA-2 generates random inferences. This finding aligns with expectations, as LLaMA-2 is primarily an English-centric model and not inherently multilingual. In an effort to address the multilingual limitations of LLaMA-2, \citet{lin2024mala500} proposed MaLA-500, a multilingual adaptation of the model that underwent fine-tuning using a causal language modeling objective. However, after this adaptation, MaLA-500 produces random-level inferences on the English subset.
    
    \item Aya-101, a 13B parameter mt5-xxl-based model~\cite{xue-etal-2021-mt5} instruction-tuned in 101 languages. It shows superior performance both in English and Croatian. 
\end{enumerate}

Based on these preliminary findings, we select the two best-performing models, Aya-101 and {BERT}i{\'c} , for further experimentation in the subsequent phases of our study. 
We report our preliminary experimental findings in \cref{tab:preliminary_experiments}. The results of our preliminary experiments are summarized in \cref{tab:preliminary_experiments}.

\begin{table*}[htpb]
\centering
\small
\begin{tabular}{lp{6cm}p{7cm}}
\toprule
\textbf{Setting} & \textbf{Description} & \textbf{Data Combination} \\
\midrule
\texttt{o} & All \textbf{o}riginal dialect-copa training data mixed together & \makecell[l]{{[}en, hr, mk, sl, sl-cer, sr, sr-tor{]}-train\\ {[}sr, sr-tor{]}-trans} \\
\midrule
\texttt{otrsl} & Combining all \textbf{o}riginal, \textbf{t}ransliterated as well as \textbf{r}everse-augmented and \textbf{s}ynthetic training data (only \textbf{l}atin script ones) & \makecell[l]{{[}en, hr, sl, sl-cer{]}-train\\ {[}sr, sr-tor, mk{]}-trans\\ {[}en, hr, mk-trans, sl, sl-cer, sr-trans, sr-tor-trans{]}-reverse\\ en-gpt4, hr-ckm-claude\\ {[}hr, sl, mk-trans, sr-trans{]}-nllb}\\
\midrule
$\texttt{otrslc}$ & Combining all original, transliterated as well as reverse-augmented and synthetic training data (Both latin and cyrillic script) & all available training data \\
\midrule
$\texttt{otrsl}_\texttt{mk-hr-ckm}$ & Selective \texttt{otrsl} setting with upsampled data count by repetition for mk, hr and hr-ckm & \makecell[l]{hr-train, mk-trans, hr-ckm-claude\\ {[}hr-train, mk-trans, hr-ckm-claude{]}-reverse\\ {[}hr, mk-trans{]}-nllb} \\
\midrule
$\texttt{otrsl}_\texttt{hr-ckm}$ & Selective \texttt{otrsl} setting with upsampled data count by repetition for hr and hr-ckm & \makecell[l]{hr-train, hr-ckm-claude\\ {[}hr-train, hr-ckm-claude{]}-reverse\\ hr-nllb} \\
\midrule
$\texttt{otrsl}_\texttt{sl-cer}$ & Same as previous but for sl and sl-cer & \makecell[l]{{[}sl, sl-cer{]}-train\\ {[}sl, sl-cer{]}-reverse\\ sl-nllb} \\
\midrule
$\texttt{otrsl}_\texttt{sr-tor}$ & Same as previous but for sr and sr-tor & \makecell[l]{{[}sr, sr-tor{]}-trans\\ {[}sr-trans, sr-tor-trans{]}-reverse\\ sr-nllb-trans} \\
\midrule
$\texttt{otrslc}_\texttt{sr-tor}$ & Same as previous but we include both transliterated as well as Cyrillic script data& \makecell[l]{{[}sr, sr-tor{]}-train, {[}sr, sr-tor{]}-trans\\ {[}sr, sr-trans, sr-tor, sr-tor-trans{]}-reverse\\ sr-nllb, sr-nllb-trans} \\
\midrule
$\texttt{otrsl}_\texttt{mix}$ & Cross-lingual mix and match using all data from \texttt{otrsl} setting & \makecell[l]{{[}en, hr, sl, sl-cer{]}-train\\ {[}sr, sr-tor, mk{]}-trans\\ {[}en, hr, mk-trans, sl, sl-cer, sr-trans, sr-tor-trans{]}-reverse\\ en-gpt4, hr-ckm-claude\\ {[}hr, sl, mk-trans, sr-trans{]}-nllb} \\
\midrule
$\texttt{otrsl}_\texttt{mix-mk-hr-ckm}$ & Cross-lingual mix and match using all data from $\texttt{otrsl}_\texttt{mk-hr-ckm}$ setting & \makecell[l]{hr-train, mk-trans, hr-ckm-claude\\ {[}hr-train, mk-trans, hr-ckm-claude{]}-reverse\\ {[}hr, mk-trans{]}-nllb} \\
\midrule
$\texttt{otrsl}_\texttt{mix-hr-ckm}$ & Cross-lingual mix and match using all data from $\texttt{otrsl}_\texttt{hr-ckm}$ setting & \makecell[l]{hr-train, hr-ckm-claude\\ {[}hr-train, hr-ckm-claude{]}-reverse\\ hr-nllb} \\
\midrule
$\texttt{otrslc}_{\texttt{mix-testset}}$ & Cross-lingual mix and match using all data from \texttt{otrslc} setting except English& all available training data except English\\
\bottomrule
\end{tabular}
\caption{After performing data augmentation, we create various data combinations by merging the augmented data blocks described in \cref{tab:data_augmentation} with the original training datasets. These carefully designed data settings are then employed to conduct task-specific fine-tuning or instruction tuning on the selected base models, enabling us to evaluate the impact of different data configurations and therefore, select the suitable ones for the test-set evaluation phase.}
\label{tab:combining_data}
\end{table*}

\paragraph{Phase 2: Data Augmentation} To address the limited size of the \task{} training dataset, which consists of only 400 instances per language, we employ various data augmentation techniques to expand the available training data. This step is crucial in mitigating the data scarcity bottleneck and improving the models' ability to generalize across diverse dialectal variations. By augmenting the training data, we aim to provide a more representative dataset for task-specific fine-tuning and instruction tuning of our selected language models. The data augmentation approaches we explore include:
\begin{enumerate}[label=\textrightarrow, leftmargin=*]
    \item The test dataset primarily contains instances written using the Latin script. Hence, we transliterate the Macedonian (mk) dataset from Cyrillic to Latin script to maintain consistency with the already available Serbian, and Torlak transliterated datasets.
    \item For each instance in the training data, we swap the premise and the correct answer choice, effectively transforming cause examples into effect examples and vice versa, thereby doubling the number of training instances. For example consider the following premise and two \texttt{`effect'} choices:

    \noindent\fbox{
\begin{minipage}{.8\columnwidth}
\small
premise: I poured water on my sleeping 
friend.

choice1: My friend awoke. \checkmark

choice2: My friend snored. $\times$
\end{minipage}}

    Now our proposed reverse-augmentation method will transform the above example in a \texttt{`casue-specific'} question as follows:

    \noindent\fbox{
\begin{minipage}{.8\columnwidth}
\small
premise: My friend awoke. 

choice1: I poured water on my sleeping 
friend.\checkmark

choice2: My friend snored. $\times$
\end{minipage}}
\\

    \item We utilize a publicly available English COPA-style synthetic dataset generated by GPT-4~\cite{Achiam2023GPT4TR}, as introduced by \citet{whitehouse-etal-2023-llm}. To expand the coverage of this synthetic data to other languages, we translate the English examples using the NLLB-6B machine translation model~\cite{nllbteam2022language} to all the four \task{} standard languages: Croatian, Macedonian, Serbian and Slovenian.
    \item The \task{} dataset does not provide any training or validation data for the Chakavian dialect. To overcome this limitation,  we compile a set of Croatian to Chakavian conversion rules and corresponding examples from online language community forums \citep{unilangCroatianchakavianFeatures}. In addition to these rules, we also gather a few Croatian to Chakavian lyrics translations~\cite{lyricstranslateElitniOdredi}. We then prompt the Claude-3 language model \citep{anthropicClaude} with these rules and examples, instructing it to translate the Croatian sentences from the \task{} training set into their Chakavian equivalents. Through this process, we create a synthetic Chakavian training set in the style of \task{}, which we refer to as \texttt{[lang]-claude}. Here is an example with ground truth Croatian to Chakavian translation (correctly translated words are bolded):

    \noindent\fbox{
\begin{minipage}{.9\columnwidth}
\small
\begin{enumerate}[label=\textrightarrow, leftmargin=*]
\item Croatian (source): Djevojka je pronašla kukca u žitaricama. Izgubila je apetit.
\item Chakavian (gold-translation): Mlada je \textbf{našla} neko blago va žitaricah. Je \textbf{zgubila} \textbf{tiek}.

\item Chakavian (claude-translation): Divojka je \textbf{našla} buba u žitarican. \textbf{Zgubila} je \textbf{tiek}.
\end{enumerate}

\end{minipage}}
\\

We observe that only a small number of words, specifically three in this instance, are correctly translated from Croatian to Chakavian. Despite the limited accuracy of the translation, this synthetic translated dataset enables us to train and evaluate models on the Chakavian dialect, despite the absence of original training data for this specific dialect. The detailed report on the dialect conversion rules and the Claude-3 prompt template used for generating the synthetic Chakavian dataset can be found in \cref{app:chak_conv}.

\end{enumerate}
\cref{tab:data_augmentation} provides a comprehensive overview of the data augmentation techniques employed and the languages covered by each approach.

\paragraph{Phase 3: Data Selection} Following the data augmentation process, we create various data combinations by merging the augmented data with the original training datasets. \cref{tab:combining_data} provides a comprehensive overview of the various training data combination settings we employ, along with their respective descriptions and the specific data sources included in each combination. These combinations are designed to investigate the impact of different data characteristics on the performance of our models. For instance, the \texttt{otrsl} setting combines all original, transliterated, reverse-augmented, and synthetic data while excluding any data written in the Cyrillic script. The rationale behind this combination is to assess whether our Latin-only \task{} test set benefits from the absence of script variations in the training data. Additionally, we introduce a language-agnostic data combination denoted as $\texttt{otrsl}_{\texttt{mix}}$, in which we perform cross-lingual modifications by ensuring that the premise, \texttt{choice1}, and \texttt{choice2} for each example are presented in different languages. This combination allows us to evaluate the models' ability to handle language-agnostic reasoning. 

\paragraph{Phase 4: Prompt Design} Encoder-based models can be fine-tuned using any of the data settings created in the previous steps. However, to perform few-shot prompting or instruction tuning with generative language models (LLMs), we need to design prompt-based instructions. During our preliminary experiments, we observed that using 4-shot same-class prompting (i.e., providing 4 cause examples for a cause-based question) yields slightly better results compared to combining 2 cause and 2 effect examples in the prompt. Specifically, this approach led to a 4.9\% improvement on the English validation set. So we opted for 4-shot same-class prompting to perform inference. 

The following prompt template is used for inference and instruction tuning of the Aya-101 model:
\\

\noindent\fbox{
\begin{minipage}
{\columnwidth}
\small
Instruction: Given the premise, \{premise\}, What is the correct \{question\} \{`before'/`after'\} this?

A: \{choice1\}

B: \{choice2\}

Correct \{question\}: \{correct\_answer\}
\end{minipage}}
\\

By designing the prompt in this manner, we provide the model with a clear instruction, the premise, and the two answer choices. The model is then expected to select the correct answer based on the given question type (cause or effect). This template is employed both during inference and instruction tuning of the Aya-101 model to ensure consistency and optimize performance on the \task{} dataset.


\paragraph{Phase 5: Task-Specific Tuning} We employ two distinct approaches for task-specific tuning of our selected models. The first approach, known as full model fine-tuning, involves updating all the weights of the model during the training process. We apply this method to the BERTi\'c model, fine-tuning it for 5-10 epochs on the \task{} dataset. However, for mid-size models like Aya-101, full fine-tuning may be unnecessarily computationally expensive, especially considering the limited amount of training data available.
To address this concern, we use LoRA (Low-Rank Adaptation) adapter tuning \citep{hu2022lora} which is a more parameter-efficient tuning approach. LoRA introduces a small number of trainable parameters in the form of low-rank matrices, which are inserted between the layers of the pre-trained model. Note that this draws from a long history of efficient adaption using dedicated units~\cite{pmlr-v97-houlsby19a,pfeiffer2020AdapterHub, faisal-anastasopoulos-2022-phylogeny}. During training, only these newly introduced parameters are updated, while the original model weights remain frozen. This approach significantly reduces the number of trainable parameters, making it more suitable for fine-tuning on smaller datasets. By employing LoRA adapter tuning, we can effectively adapt the Aya-101 model to the \task{} dataset without the need for full model fine-tuning, thereby striking a balance between performance and computational efficiency.

\section{Results and Discussion}
\label{sec:result}
In this section, we present and discuss the results of our experiments on the \task{} dataset.

\begin{table*}[htpb]
\centering
\small
\begin{tabular}{lllrrrrrrr}
\toprule
Base Model & Setting & en & hr & mk & sl & sl-cer & sr & sr-tor & Avg. (acc) \\
\midrule
\multicolumn{10}{c}{Takeaway 1: Combining all training data helps} \\
\cmidrule{1-10}
{BERT}i{\'c} & Finetune (hr) &-  & 64 &-  & - & - & - &-  & - \\
{BERT}i{\'c} & \texttt{o} & 0.65 & 0.67 & 0.55 & 0.67 & 0.49 & 0.66 & 0.59 & 0.61 \\
\midrule\\
\multicolumn{10}{c}{Takeaway 2: Data augmentation helps for low-resource languages in most cases} \\
\cmidrule{1-10}
{BERT}i{\'c} & $\texttt{otrslc}$ & 0.46 & 0.7 & 0.69 & 0.67 & 0.42 & 0.68 & 0.65 & 0.61 \\
\midrule\\
\multicolumn{10}{c}{Takeaway 3: Script choice makes a difference (Using only Latin script performs better on Latin script evaluation)} \\
\cmidrule{1-10}
{BERT}i{\'c} & $\texttt{otrsl}$ & 0.51 & 0.77 & 0.64 & 0.64 & 0.59 & 0.72 & 0.64 & 0.65 \\
\midrule\\
\multicolumn{10}{c}{Takeaway 4: Cross-lingual mix-and-match effect: inconclusive} \\
\cmidrule{1-10}
{BERT}i{\'c} & $\texttt{otrsl}_\texttt{mix}$ & 0.56 & 0.76 & 0.68 & 0.57 & 0.52 & 0.66 & 0.63 & 0.63 \\
\midrule\\
\multicolumn{10}{c}{Takeaway 5: Upsampling certain language groups might help targeted evaluation in some cases} \\
\cmidrule{1-10}
{BERT}i{\'c} & $\texttt{otrsl}_\texttt{sr-tor}$ & 0.55 & 0.74 & 0.63 & 0.61 & 0.48 & 0.64 & 0.66 & 0.62 \\
{BERT}i{\'c} & $\texttt{otrsl}_\texttt{sl-cer}$ & 0.54 & 0.68 & 0.65 & 0.57 & 0.58 & 0.66 & 0.64 & 0.62 \\
\midrule\\
\multicolumn{10}{c}{Takeaway 6: Instruction tuning helps} \\
\cmidrule{1-10}
Aya-101 & 4-shot & 0.83 & 0.77 & 0.75 & 0.76 & 0.62 & 0.81 & 0.73 & 0.75 \\
\midrule\\
\multicolumn{10}{c}{Takeaway 7: Further task-specific instruction tuning helps even more} \\
\cmidrule{1-10}
Aya-101 & $\texttt{otrsl}$ & 0.86 & 0.79 & 0.81 & 0.91 & 0.7 & 0.82 & 0.77 & 0.81 \\
\midrule\\
\end{tabular}
\begin{tabular}{llrrrrrrrrr}
\multicolumn{10}{c}{Takeaway 7: Training for 10 epochs instead of 5: inconclusive} \\
\cmidrule{1-11}
 setting & epochs & en & hr & mk & sl & sl-cer & sr & sr-tor & mean & max count \\ \midrule
 $\texttt{otrsl}$ & 10 & \underline{0.52} & \underline{0.74} & \underline{0.65} & 0.56 & \underline{0.61} & \underline{0.67} & 0.59 & \underline{0.62} & 5 \\
 & 5 & 0.50 & 0.72 & 0.62 & \underline{0.57} & 0.58 & 0.65 & \underline{0.61} & 0.61 & 2 \\ \midrule
$\texttt{otrsl}_\texttt{mk-hr-ckm}$ & 10 & 0.49 & 0.76 & 0.66 & 0.62 & 0.51 & 0.67 & \underline{0.70} & 0.63 & 1 \\
 & 5 & 0.48 & 0.76 & \underline{0.68} & \underline{0.64} & \underline{0.52} & \underline{0.71} & 0.67 & \underline{0.64} & 4 \\ \midrule
$\texttt{otrsl}_\texttt{sl-cer}$ & 10 & \underline{0.56} & 0.64 & \underline{0.65} & \underline{0.62} & \underline{0.58} & \underline{0.65} & 0.60 & 0.62 & 5 \\
 & 5 & 0.54 & \underline{0.68} & \underline{0.65} & 0.57 & \underline{0.58} & \underline{0.66} & \underline{0.64} & 0.62 & 5 \\ \midrule
$\texttt{otrslc}_\texttt{sr-tor}$ & 10 & \underline{0.50} & \underline{0.70} & \underline{0.65} & 0.55 & 0.48 & 0.62 & \underline{0.68} & 0.60 & 4 \\
 & 5 & 0.49 & 0.69 & 0.63 & \underline{0.60} & \underline{0.51} & \underline{0.62} & \underline{0.70} & \underline{0.61} & 4 \\ 
\bottomrule
\end{tabular}
\caption{Takeaways from incremental experiments performed on the \task{} validation dataset. The best language-specific scores for each setting are underlined (Takeaway 7).}
\label{tab:takeaway}
\end{table*}

\begin{table*}[htbp]
\centering
\small
\begin{tabular}{l|lp{8cm}|lll|l}
\toprule
 Team & Base Model&System Description & sl-cer & hr-ckm & sr-tor & Avg. (acc) \\
\midrule
\multicolumn{7}{c}{Closed Source Model Weights}\\ 
\midrule
 JSI    & GPT-4& 10-shot with first 10 test instances (without answer) & \textbf{0.734} & \textbf{0.890} & \textbf{0.974} & \textbf{0.866} \\
 UNIRI  & GPT-4& RAG implementation; Chakavian and Cerkno lexical dictionary; Reasoning instruction and self referral grading task & 0.708 & 0.764 & - & - \\
 UNIRI  & GPT-4& 0-shot iterative prompt          & 0.664 & 0.774 & 0.894 & 0.777 \\
\midrule
\multicolumn{7}{c}{Open Source Model Weights}\\ 
\midrule
 GmuNLP & Aya-101& 4-shot prompting                & 0.694 & 0.756 & 0.840 & \textbf{0.763} \\
  GmuNLP & Aya-101& LORA adapter tuning on $\texttt{otrsl}_{\texttt{hr-ckm}}\rightarrow$ 4-shot prompting & \textbf{0.700} & 0.750 & 0.824 & 0.758  \\
 GmuNLP & Aya-101& LORA adapter tuning on $\texttt{otrsl}\rightarrow$ 4-shot prompting          & 0.682 & \textbf{0.760} & 0.824 & 0.755 \\
 GmuNLP & Aya-101& LORA adapter tuning on $\texttt{otrsl}_{\texttt{hr-ckm}}\rightarrow$ 4-shot prompting & 0.660 & 0.742 & \textbf{0.848} & 0.750 \\
 WueNLP & Mixtral& LORA adapter tuning on standard variety of target dialect & 0.556 & 0.606 & 0.738 & 0.633 \\
  CLaC   & XLM-R& Fine-tuning XLM-RoBERTa base for multiple choice QA task        & 0.564 & 0.522 & 0.570 & 0.552 \\
\bottomrule
\end{tabular}
\caption{Performance comparison of different submissions on \task{} test set.}
\label{tab:all_team}
\end{table*}

\begin{table*}[h]
\small
\centering
\begin{tabular}{llrrrrr}
\toprule
base model & setting & epoch & sl-cer & hr-ckm & sr-tor & Avg. (acc) \\ \midrule
Aya-101 & 4-shot & - & 0.694 & \textbf{0.756} & 0.840 & \textbf{0.763} \\
Aya-101 & $\texttt{otrslc}_\texttt{mix-mk-hr-ckm}$ & 5 & 0.690 & \textbf{0.756} & 0.836 & 0.761 \\
Aya-101 &  $\texttt{otrsl}_\texttt{mix-hr-ckm}$ & 5 & \textbf{0.700} & 0.750 & 0.824 & 0.758 \\
Aya-101 &  \texttt{otrsl} & 5 &0.682 & \textbf{0.760} & 0.824 & 0.755 \\
Aya-101 &  $\texttt{otrsl}_\texttt{mk-hr-ckm}$ &5 & 0.660 & 0.742 & 0.848 & 0.750 \\
Aya-101 &  $\texttt{otrsl}_\texttt{sl-cer}$ & 5& 0.686 & 0.718 & 0.836 & 0.747 \\
\midrule
{BERT}i{\'c} &  $\texttt{otrsl}_\texttt{hr-ckm} $& 10 & 0.572 & 0.626 & \textbf{0.722} & \textbf{0.640} \\
{BERT}i{\'c} &  $\texttt{otrsl}_\texttt{mk-hr-ckm}$ & 5 & \textbf{0.582} & \textbf{0.634} & 0.682 & 0.633 \\
{BERT}i{\'c} &  $\texttt{otrslc}_\texttt{mix-testset}$ &5& 0.576 & 0.622 & 0.692 & 0.630 \\
{BERT}i{\'c} &  \texttt{otrsl}&10 & 0.540 & 0.622 & 0.700 & 0.621 \\
\bottomrule
\end{tabular}
\caption{GMUNLP system submissions for test-set evaluation. The best dialect-specific scores for each base-model type are bolded.}
\label{tab:gmunlp}
\end{table*}
\subsection{Validation Set Insights}

\cref{tab:takeaway} summarizes the key takeaways from our incremental experiments conducted on the validation dataset using {BERT}i{\'c} and Aya-101. First, we observe that combining datasets from multiple languages boosts performance on the Croatian subset, as opposed to training on a single language. This finding motivates us to utilize all available training data and prepare different data combinations, as described in \cref{tab:combining_data}.
Second, we find that increasing the data quantity through various data augmentation techniques (\cref{tab:data_augmentation}) primarily improves performance for most languages and low-resource dialects. Furthermore, discarding instances written in the Cyrillic script can boost performance for certain languages and dialects (e.g., Croatian, Cerkno, and Serbian), while hurting others.

We also explore cross-lingual mix-and-match strategies, but we do not find any conclusive patterns indicating that this approach consistently makes the model more language-agnostic, as it helps in some cases while hindering performance in others. Additionally, we experiment with discarding English examples and upsampling specific language groups (e.g., Serbian and Torlak examples for the Torlak dialect), which leads to slight performance improvements for the Torlak dialect.

Notably, we observe that full fine-tuning of the comparatively smaller, non-instruction-tuned, but language-specific {BERT}i{\'c} model cannot surpass the performance of the multilingual, instruction-tuned Aya-101 model. Finally, we apply the same data combinations to perform instruction tuning on the Aya-101 model and observe an overall performance boost. However, our experiments with different numbers of training epochs (5 and 10) yield inconclusive findings.

\subsection{Test Set Insights}

\paragraph{Team-specific ranking} \Cref{tab:all_team} presents a comparison of the best-performing submissions from different teams on the \task{} test set. We categorize the submissions into two groups: Category 1 includes teams that utilize closed-source model weights, while Category 2 consists of teams that rely on open-source model weights. Our submissions belong to the latter category. We observe that the closed-source GPT-4 model achieves the best overall performance. Team JSI employs GPT-4 with a 10-shot prompting approach, where they provide the first 10 test instances without revealing the answers. Interestingly, even the 0-shot prompting using GPT-4 (by team UNRI) outperforms all submissions in Category 2 using open-source models. Among the Category 2 submissions, GMUNLP (our submission) achieves the highest performance on all varieties. The base Aya-101 model with 4-shot prompting yields the best average score across all languages. However, LoRA adapter tuning on different data combinations results in language-specific best scores.

\paragraph{\gmnlp{} submission} \Cref{tab:gmunlp} presents the results of our selected 10 system submissions. We observe that the best performance achieved by the {BERT}i{\'c} model on the \texttt{otrsl} setting is 62\%, which is approximately 17\% lower compared to the \texttt{otrcl}-tuned Aya-101 model. When comparing language-specific results, we find that the Torlak (sr-tor) dialect is the easiest to predict for both the Aya-101 and BERTi'c models, while the Cerkno dialect (sl-cer) proves to be the most challenging to learn. 

Interestingly, upsampling the Cerkno dialect-related data ($\texttt{otrsl}_{\texttt{sl-cer}}$-tuned) does not yield the best score for the Cerkno test-set. Instead, upsampling the Chakavian dialect-related data using the $\texttt{otrsl}_{\texttt{mk-hr-ckm}}$ setting leads to better scores on the Cerkno test set. This observation holds true for both the Aya-101 and {BERT}i{\'c} base models, indicating that leveraging data from more closely related languages does not always provide the most significant benefit. We believe this phenomenon warrants further investigation to gain a deeper understanding of the complex interplay between language-relatedness and task-specific model performance.


\section{Conclusion}
\label{sec:conclusion}
In this study, we explored the impact of data augmentation techniques on fine-tuning multilingual language models for improving common sense reasoning in dialectal variations. Our experiments encompassed a range of language models, from smaller to mid-sized architectures, to investigate their adaptability to dialectal nuances. The observed variations in performance and the upper limits achieved by different models reflect the diverse ways in which language models handle and adapt to dialectal variations. The insights gained from this work may contribute to the development of more robust and adaptable language models that can handle the challenges posed by dialectal variations. Future work can explore advanced data augmentation techniques, investigate the impact of domain-specific knowledge integration, and develop novel architectures tailored to the unique characteristics of dialects.

\section*{Acknowledgements}
This work has been generously supported by the National Science Foundation under grants IIS-2125466 and IIS-2327143.
We are thankful to the anonymous reviewers and area chairs for their constructive feedback.
This project was supported by resources provided by the Office of Research Computing at George Mason University (\url{https://orc.gmu.edu}) and funded in part by grants from the National Science Foundation (Awards Number 1625039 and 2018631).

\bibliography{anthology,custom}
\bibliographystyle{acl_natbib}

\appendix

\section{Croatian to Chakavian conversion rules}
\label{app:chak_conv}
\input{chaka_conv}

\end{document}

%% file: chaka_conv.tex
Here we report the collected Croatian to Chakavian conversion rules and their corresponding examples~\cite{unilangCroatianchakavianFeatures}, the lyrics translations~\cite{lyricstranslateElitniOdredi} and the Claude-3 prompt we used to generate the Cahakavian synthetic translations.

\paragraph{Conversion rules}
\begin{verbatim}
Rule: m's at the end of words become n's
Examples:
Ja sam = Ja san
osam = osan
s ženom = s ženon
vidim = vidin

Rule: đ becomes j
Examples:
mlađi=>mlaji
među=>meju

Rule: The sounds ć and đ actually do not 
exist separately from č and dž

Rule: genitive plural in the feminine and 
neuter takes a zero ending: 
Examples: žena = žen, sela = sel

Rule: dative/instrumental plurals are 
also somewhat different.

Rule: Masculine and neuter nouns can have
an alternate ending.
Examples: 
it can be gradovima or gradoviman

Rule: For feminine nouns, it's shortened. 
Examples:
ženama=>ženan

Rule:-ao endings are shortened to just -a
Examples:
išao je => iša je
rekao sam => reka san

Rule:-io endings change to -ija
Examples:
govorio je => govorija je
vidio sam => vidija san

Rule:ča, aside from meaning "what" can 
also be used as a particle meaning "away" 
or "out".
Examples:
gremo ča (let's get out of here)

Rule: some of the third person plural 
forms can often be extended into longer 
-u ending forms.
Examples:
govore -> govoru -> govoridu
rade -> radu -> radidu
pišu -> pišedu

Rule: Infinitive is shortened .
There is no I at the end.
In some speeches there is neither T/Ć at 
the end
Examples: 
bit =>bi
pivat =>piva
znat => zna
plivat => pliva
ronit => roni

Rule: change lj=>j
Examples:
zaljubiti se=>zajubit se
ljubav =>jubav
ljuska =>juska
ljudi=>judi

Rule: Change O => E in some cases
Example:
nekoga-nikega
svakoga-svakega
tomu-temu
toga-tega
bijeloga-bilega
jednoga-jenega
jednomu-jenemu

Rule: In standard croatian third person 
plural has ending in some verbs E. 
In chakavian it is always U
Examples:
vide =>vidu
hoće =>hoću
stoje =>stoju
stave =>stavu
motre =>motru
leže=>ležu

Rule: Change H=>V
Examples:
kruh =>kruv
kuhati =>kuvati
suh =>suv
gluh =>gluv[/b]

\end{verbatim}

\paragraph{Lyrics translations} We collected Croatian to Chakavian lyrics translations from the \url{lyricstranslate.com} site~\cite{lyricstranslateElitniOdredi}.

\paragraph{Claude-3 prompt} We use the following prompt consisting the the above mentioned conversion rules, examples and lyrics translations:

\begin{verbatim}
<conversion_file.txt>

This file contains Croatian to Chakavian 
dialect conversion grammar rules with 
examples.

Now here are some Croatian sentences and 
it's parallel Chakavian sentences: 

<Croatian_lyrics.txt> 
<Chakavian_lyrics.txt>. 

Given these resources, I want you to translate 
the following Croatian sentences to Chakavian 
dialect.

<Croatian_training_sentences.txt>
\end{verbatim}

%% file: acl2023.bbl
\begin{thebibliography}{21}
\expandafter\ifx\csname natexlab\endcsname\relax\def\natexlab#1{#1}\fi

\bibitem[{ant()}]{anthropicClaude}

\newblock {C}laude --- anthropic.com.
\newblock \url{https://www.anthropic.com/claude}.
\newblock [Accessed 25-03-2024].

\bibitem[{uni()}]{unilangCroatianchakavianFeatures}

\newblock {C}roatian-chakavian features -{U}ni{L}ang --- forum.unilang.org.
\newblock \url{https://forum.unilang.org/viewtopic.php?t=14771}.
\newblock [Accessed 28-03-2024].

\bibitem[{lyr()}]{lyricstranslateElitniOdredi}

\newblock {E}litni {O}dredi - {L}jubavi {M}oja lyrics + {C}roatian ({C}hakavian dialect) translation --- lyricstranslate.com.
\newblock \url{https://lyricstranslate.com/en/ljubavi-moja-jubavi-moja.html}.
\newblock [Accessed 28-03-2024].

\bibitem[{Achiam et~al.(2023)Achiam, Adler, Agarwal, Ahmad, Akkaya, Aleman, Almeida, Altenschmidt, Altman, Anadkat, Avila, Babuschkin, Balaji, Balcom, Baltescu, Bao, Bavarian, Belgum, Bello, Berdine, Bernadett-Shapiro, Berner, Bogdonoff, Boiko, Boyd, Brakman, Brockman, Brooks, Brundage, Button, Cai, Campbell, Cann, Carey, Carlson, Carmichael, Chan, Chang, Chantzis, Chen, Chen, Chen, Chen, Chen, Chess, Cho, Chu, Chung, Cummings, Currier, Dai, Decareaux, Degry, Deutsch, Deville, Dhar, Dohan, Dowling, Dunning, Ecoffet, Eleti, Eloundou, Farhi, Fedus, Felix, Fishman, Forte, Fulford, Gao, Georges, Gibson, Goel, Gogineni, Goh, Gontijo-Lopes, Gordon, Grafstein, Gray, Greene, Gross, Gu, Guo, Hallacy, Han, Harris, He, Heaton, Heidecke, Hesse, Hickey, Hickey, Hoeschele, Houghton, Hsu, Hu, Hu, Huizinga, Jain, Jain, Jang, Jiang, Jiang, Jin, Jin, Jomoto, Jonn, Jun, Kaftan, Kaiser, Kamali, Kanitscheider, Keskar, Khan, Kilpatrick, Kim, Kim, Kim, Kirchner, Kiros, Knight, Kokotajlo, Kondraciuk, Kondrich, Konstantinidis, Kosic,
  Krueger, Kuo, Lampe, Lan, Lee, Leike, Leung, Levy, Li, Lim, Lin, Lin, Litwin, Lopez, Lowe, Lue, Makanju, Malfacini, Manning, Markov, Markovski, Martin, Mayer, Mayne, McGrew, McKinney, McLeavey, McMillan, McNeil, Medina, Mehta, Menick, Metz, Mishchenko, Mishkin, Monaco, Morikawa, Mossing, Mu, Murati, Murk, M'ely, Nair, Nakano, Nayak, Neelakantan, Ngo, Noh, Long, O'Keefe, Pachocki, Paino, Palermo, Pantuliano, Parascandolo, Parish, Parparita, Passos, Pavlov, Peng, Perelman, de~Avila Belbute~Peres, Petrov, de~Oliveira~Pinto, Pokorny, Pokrass, Pong, Powell, Power, Power, Proehl, Puri, Radford, Rae, Ramesh, Raymond, Real, Rimbach, Ross, Rotsted, Roussez, Ryder, Saltarelli, Sanders, Santurkar, Sastry, Schmidt, Schnurr, Schulman, Selsam, Sheppard, Sherbakov, Shieh, Shoker, Shyam, Sidor, Sigler, Simens, Sitkin, Slama, Sohl, Sokolowsky, Song, Staudacher, Such, Summers, Sutskever, Tang, Tezak, Thompson, Tillet, Tootoonchian, Tseng, Tuggle, Turley, Tworek, Uribe, Vallone, Vijayvergiya, Voss, Wainwright, Wang, Wang,
  Wang, Ward, Wei, Weinmann, Welihinda, Welinder, Weng, Weng, Wiethoff, Willner, Winter, Wolrich, Wong, Workman, Wu, Wu, Wu, Xiao, Xu, Yoo, Yu, Yuan, Zaremba, Zellers, Zhang, Zhang, Zhao, Zheng, Zhuang, Zhuk, and Zoph}]{Achiam2023GPT4TR}
OpenAI~Josh Achiam, Steven Adler, Sandhini Agarwal, Lama Ahmad, Ilge Akkaya, Florencia~Leoni Aleman, Diogo Almeida, Janko Altenschmidt, Sam Altman, Shyamal Anadkat, Red Avila, Igor Babuschkin, Suchir Balaji, Valerie Balcom, Paul Baltescu, Haiming Bao, Mo~Bavarian, Jeff Belgum, Irwan Bello, Jake Berdine, Gabriel Bernadett-Shapiro, Christopher Berner, Lenny Bogdonoff, Oleg Boiko, Madelaine Boyd, Anna-Luisa Brakman, Greg Brockman, Tim Brooks, Miles Brundage, Kevin Button, Trevor Cai, Rosie Campbell, Andrew Cann, Brittany Carey, Chelsea Carlson, Rory Carmichael, Brooke Chan, Che Chang, Fotis Chantzis, Derek Chen, Sully Chen, Ruby Chen, Jason Chen, Mark Chen, Benjamin Chess, Chester Cho, Casey Chu, Hyung~Won Chung, Dave Cummings, Jeremiah Currier, Yunxing Dai, Cory Decareaux, Thomas Degry, Noah Deutsch, Damien Deville, Arka Dhar, David Dohan, Steve Dowling, Sheila Dunning, Adrien Ecoffet, Atty Eleti, Tyna Eloundou, David Farhi, Liam Fedus, Niko Felix, Sim'on~Posada Fishman, Juston Forte, Isabella Fulford, Leo Gao,
  Elie Georges, Christian Gibson, Vik Goel, Tarun Gogineni, Gabriel Goh, Raphael Gontijo-Lopes, Jonathan Gordon, Morgan Grafstein, Scott Gray, Ryan Greene, Joshua Gross, Shixiang~Shane Gu, Yufei Guo, Chris Hallacy, Jesse Han, Jeff Harris, Yuchen He, Mike Heaton, Johannes Heidecke, Chris Hesse, Alan Hickey, Wade Hickey, Peter Hoeschele, Brandon Houghton, Kenny Hsu, Shengli Hu, Xin Hu, Joost Huizinga, Shantanu Jain, Shawn Jain, Joanne Jang, Angela Jiang, Roger Jiang, Haozhun Jin, Denny Jin, Shino Jomoto, Billie Jonn, Heewoo Jun, Tomer Kaftan, Lukasz Kaiser, Ali Kamali, Ingmar Kanitscheider, Nitish~Shirish Keskar, Tabarak Khan, Logan Kilpatrick, Jong~Wook Kim, Christina Kim, Yongjik Kim, Hendrik Kirchner, Jamie~Ryan Kiros, Matthew Knight, Daniel Kokotajlo, Lukasz Kondraciuk, Andrew Kondrich, Aris Konstantinidis, Kyle Kosic, Gretchen Krueger, Vishal Kuo, Michael Lampe, Ikai Lan, Teddy Lee, Jan Leike, Jade Leung, Daniel Levy, Chak~Ming Li, Rachel Lim, Molly Lin, Stephanie Lin, Mateusz Litwin, Theresa Lopez, Ryan
  Lowe, Patricia Lue, Anna~Adeola Makanju, Kim Malfacini, Sam Manning, Todor Markov, Yaniv Markovski, Bianca Martin, Katie Mayer, Andrew Mayne, Bob McGrew, Scott~Mayer McKinney, Christine McLeavey, Paul McMillan, Jake McNeil, David Medina, Aalok Mehta, Jacob Menick, Luke Metz, Andrey Mishchenko, Pamela Mishkin, Vinnie Monaco, Evan Morikawa, Daniel~P. Mossing, Tong Mu, Mira Murati, Oleg Murk, David M'ely, Ashvin Nair, Reiichiro Nakano, Rajeev Nayak, Arvind Neelakantan, Richard Ngo, Hyeonwoo Noh, Ouyang Long, Cullen O'Keefe, Jakub~W. Pachocki, Alex Paino, Joe Palermo, Ashley Pantuliano, Giambattista Parascandolo, Joel Parish, Emy Parparita, Alexandre Passos, Mikhail Pavlov, Andrew Peng, Adam Perelman, Filipe de~Avila Belbute~Peres, Michael Petrov, Henrique~Pond{\'e} de~Oliveira~Pinto, Michael Pokorny, Michelle Pokrass, Vitchyr~H. Pong, Tolly Powell, Alethea Power, Boris Power, Elizabeth Proehl, Raul Puri, Alec Radford, Jack Rae, Aditya Ramesh, Cameron Raymond, Francis Real, Kendra Rimbach, Carl Ross, Bob
  Rotsted, Henri Roussez, Nick Ryder, Mario~D. Saltarelli, Ted Sanders, Shibani Santurkar, Girish Sastry, Heather Schmidt, David Schnurr, John Schulman, Daniel Selsam, Kyla Sheppard, Toki Sherbakov, Jessica Shieh, Sarah Shoker, Pranav Shyam, Szymon Sidor, Eric Sigler, Maddie Simens, Jordan Sitkin, Katarina Slama, Ian Sohl, Benjamin~D. Sokolowsky, Yang Song, Natalie Staudacher, Felipe~Petroski Such, Natalie Summers, Ilya Sutskever, Jie Tang, Nikolas~A. Tezak, Madeleine Thompson, Phil Tillet, Amin Tootoonchian, Elizabeth Tseng, Preston Tuggle, Nick Turley, Jerry Tworek, Juan Felipe~Cer'on Uribe, Andrea Vallone, Arun Vijayvergiya, Chelsea Voss, Carroll Wainwright, Justin~Jay Wang, Alvin Wang, Ben Wang, Jonathan Ward, Jason Wei, CJ~Weinmann, Akila Welihinda, Peter Welinder, Jiayi Weng, Lilian Weng, Matt Wiethoff, Dave Willner, Clemens Winter, Samuel Wolrich, Hannah Wong, Lauren Workman, Sherwin Wu, Jeff Wu, Michael Wu, Kai Xiao, Tao Xu, Sarah Yoo, Kevin Yu, Qiming Yuan, Wojciech Zaremba, Rowan Zellers, Chong
  Zhang, Marvin Zhang, Shengjia Zhao, Tianhao Zheng, Juntang Zhuang, William Zhuk, and Barret Zoph. 2023.
\newblock \href {https://api.semanticscholar.org/CorpusID:257532815} {Gpt-4 technical report}.

\bibitem[{Chifu et~al.(2024)Chifu, Glava\v{s}, Ionescu, Ljube\v{s}i\'{c}, Mileti\'{c}, Mileti\'{c}, Scherrer, and Vuli\'{c}}]{chifu-etal-2024-vardial}
Adrian Chifu, Goran Glava\v{s}, Radu Ionescu, Nikola Ljube\v{s}i\'{c}, Aleksandra Mileti\'{c}, Filip Mileti\'{c}, Yves Scherrer, and Ivan Vuli\'{c}. 2024.
\newblock Var{D}ial evaluation campaign 2024: Commonsense reasoning in dialects and multi-label similar language identification.
\newblock In \emph{Eleventh Workshop on NLP for Similar Languages, Varieties and Dialects (VarDial 2024)}, Mexico City, Mexico. Association for Computational Linguistics.

\bibitem[{Conneau et~al.(2020)Conneau, Khandelwal, Goyal, Chaudhary, Wenzek, Guzm{\'a}n, Grave, Ott, Zettlemoyer, and Stoyanov}]{conneau-etal-2020-unsupervised}
Alexis Conneau, Kartikay Khandelwal, Naman Goyal, Vishrav Chaudhary, Guillaume Wenzek, Francisco Guzm{\'a}n, Edouard Grave, Myle Ott, Luke Zettlemoyer, and Veselin Stoyanov. 2020.
\newblock \href {https://doi.org/10.18653/v1/2020.acl-main.747} {Unsupervised cross-lingual representation learning at scale}.
\newblock In \emph{Proceedings of the 58th Annual Meeting of the Association for Computational Linguistics}, pages 8440--8451, Online. Association for Computational Linguistics.

\bibitem[{Devlin et~al.(2019)Devlin, Chang, Lee, and Toutanova}]{devlin-etal-2019-bert}
Jacob Devlin, Ming-Wei Chang, Kenton Lee, and Kristina Toutanova. 2019.
\newblock \href {https://doi.org/10.18653/v1/N19-1423} {{BERT}: Pre-training of deep bidirectional transformers for language understanding}.
\newblock In \emph{Proceedings of the 2019 Conference of the North {A}merican Chapter of the Association for Computational Linguistics: Human Language Technologies, Volume 1 (Long and Short Papers)}, pages 4171--4186, Minneapolis, Minnesota. Association for Computational Linguistics.

\bibitem[{Faisal et~al.(2024)Faisal, Ahia, Srivastava, Ahuja, Chiang, Tsvetkov, and Anastasopoulos}]{faisal2024dialectbench}
Fahim Faisal, Orevaoghene Ahia, Aarohi Srivastava, Kabir Ahuja, David Chiang, Yulia Tsvetkov, and Antonios Anastasopoulos. 2024.
\newblock \href {http://arxiv.org/abs/2403.11009} {Dialectbench: A nlp benchmark for dialects, varieties, and closely-related languages}.

\bibitem[{Faisal and Anastasopoulos(2022)}]{faisal-anastasopoulos-2022-phylogeny}
Fahim Faisal and Antonios Anastasopoulos. 2022.
\newblock \href {https://aclanthology.org/2022.aacl-main.34} {Phylogeny-inspired adaptation of multilingual models to new languages}.
\newblock In \emph{Proceedings of the 2nd Conference of the Asia-Pacific Chapter of the Association for Computational Linguistics and the 12th International Joint Conference on Natural Language Processing (Volume 1: Long Papers)}, pages 434--452, Online only. Association for Computational Linguistics.

\bibitem[{Gordon et~al.(2011)Gordon, Kozareva, and Roemmele}]{Gordon2011ChoiceOP}
Andrew~S. Gordon, Zornitsa Kozareva, and Melissa Roemmele. 2011.
\newblock \href {https://api.semanticscholar.org/CorpusID:434646} {Choice of plausible alternatives: An evaluation of commonsense causal reasoning}.
\newblock In \emph{AAAI Spring Symposium: Logical Formalizations of Commonsense Reasoning}.

\bibitem[{Houlsby et~al.(2019)Houlsby, Giurgiu, Jastrzebski, Morrone, De~Laroussilhe, Gesmundo, Attariyan, and Gelly}]{pmlr-v97-houlsby19a}
Neil Houlsby, Andrei Giurgiu, Stanislaw Jastrzebski, Bruna Morrone, Quentin De~Laroussilhe, Andrea Gesmundo, Mona Attariyan, and Sylvain Gelly. 2019.
\newblock \href {https://proceedings.mlr.press/v97/houlsby19a.html} {Parameter-efficient transfer learning for {NLP}}.
\newblock In \emph{Proceedings of the 36th International Conference on Machine Learning}, volume~97 of \emph{Proceedings of Machine Learning Research}, pages 2790--2799. PMLR.

\bibitem[{Hu et~al.(2022)Hu, Shen, Wallis, Allen-Zhu, Li, Wang, Wang, and Chen}]{hu2022lora}
Edward~J Hu, Yelong Shen, Phillip Wallis, Zeyuan Allen-Zhu, Yuanzhi Li, Shean Wang, Lu~Wang, and Weizhu Chen. 2022.
\newblock \href {https://openreview.net/forum?id=nZeVKeeFYf9} {Lo{RA}: Low-rank adaptation of large language models}.
\newblock In \emph{International Conference on Learning Representations}.

\bibitem[{Lin et~al.(2024)Lin, Ji, Tiedemann, Martins, and Schütze}]{lin2024mala500}
Peiqin Lin, Shaoxiong Ji, Jörg Tiedemann, André F.~T. Martins, and Hinrich Schütze. 2024.
\newblock \href {http://arxiv.org/abs/2401.13303} {Mala-500: Massive language adaptation of large language models}.

\bibitem[{Ljube{\v{s}}i{\'c} and Lauc(2021)}]{ljubesic-lauc-2021-bertic}
Nikola Ljube{\v{s}}i{\'c} and Davor Lauc. 2021.
\newblock \href {https://aclanthology.org/2021.bsnlp-1.5} {{BERT}i{\'c} - the transformer language model for {B}osnian, {C}roatian, {M}ontenegrin and {S}erbian}.
\newblock In \emph{Proceedings of the 8th Workshop on Balto-Slavic Natural Language Processing}, pages 37--42, Kiyv, Ukraine. Association for Computational Linguistics.

\bibitem[{Ljube\v{s}i\'{c} et~al.(2024)Ljube\v{s}i\'{c}, Galant, Ben\v{c}ina, \v{C}ibej, Milosavljevi\'{c}, Rupnik, and Kuzman}]{ljubesic-etal-2024-dialect}
Nikola Ljube\v{s}i\'{c}, Nada Galant, Sonja Ben\v{c}ina, Jaka \v{C}ibej, Stefan Milosavljevi\'{c}, Peter Rupnik, and Taja Kuzman. 2024.
\newblock {DIALECT-COPA}: Extending the standard translations of the {COPA} causal commonsense reasoning dataset to south slavic dialects.
\newblock In \emph{Eleventh Workshop on NLP for Similar Languages, Varieties and Dialects (VarDial 2024)}, Mexico City, Mexico. Association for Computational Linguistics.

\bibitem[{Pfeiffer et~al.(2020)Pfeiffer, R\"uckl\'{e}, Poth, Kamath, Vuli\'{c}, Ruder, Cho, and Gurevych}]{pfeiffer2020AdapterHub}
Jonas Pfeiffer, Andreas R\"uckl\'{e}, Clifton Poth, Aishwarya Kamath, Ivan Vuli\'{c}, Sebastian Ruder, Kyunghyun Cho, and Iryna Gurevych. 2020.
\newblock \href {https://www.aclweb.org/anthology/2020.emnlp-demos.7} {Adapterhub: A framework for adapting transformers}.
\newblock In \emph{Proceedings of the 2020 Conference on Empirical Methods in Natural Language Processing (EMNLP 2020): Systems Demonstrations}, pages 46--54, Online. Association for Computational Linguistics.

\bibitem[{Team et~al.(2022)Team, Costa-jussà, Cross, Çelebi, Elbayad, Heafield, Heffernan, Kalbassi, Lam, Licht, Maillard, Sun, Wang, Wenzek, Youngblood, Akula, Barrault, Gonzalez, Hansanti, Hoffman, Jarrett, Sadagopan, Rowe, Spruit, Tran, Andrews, Ayan, Bhosale, Edunov, Fan, Gao, Goswami, Guzmán, Koehn, Mourachko, Ropers, Saleem, Schwenk, and Wang}]{nllbteam2022language}
NLLB Team, Marta~R. Costa-jussà, James Cross, Onur Çelebi, Maha Elbayad, Kenneth Heafield, Kevin Heffernan, Elahe Kalbassi, Janice Lam, Daniel Licht, Jean Maillard, Anna Sun, Skyler Wang, Guillaume Wenzek, Al~Youngblood, Bapi Akula, Loic Barrault, Gabriel~Mejia Gonzalez, Prangthip Hansanti, John Hoffman, Semarley Jarrett, Kaushik~Ram Sadagopan, Dirk Rowe, Shannon Spruit, Chau Tran, Pierre Andrews, Necip~Fazil Ayan, Shruti Bhosale, Sergey Edunov, Angela Fan, Cynthia Gao, Vedanuj Goswami, Francisco Guzmán, Philipp Koehn, Alexandre Mourachko, Christophe Ropers, Safiyyah Saleem, Holger Schwenk, and Jeff Wang. 2022.
\newblock \href {http://arxiv.org/abs/2207.04672} {No language left behind: Scaling human-centered machine translation}.

\bibitem[{Touvron et~al.(2023)Touvron, Martin, Stone, Albert, Almahairi, Babaei, Bashlykov, Batra, Bhargava, Bhosale, Bikel, Blecher, Ferrer, Chen, Cucurull, Esiobu, Fernandes, Fu, Fu, Fuller, Gao, Goswami, Goyal, Hartshorn, Hosseini, Hou, Inan, Kardas, Kerkez, Khabsa, Kloumann, Korenev, Koura, Lachaux, Lavril, Lee, Liskovich, Lu, Mao, Martinet, Mihaylov, Mishra, Molybog, Nie, Poulton, Reizenstein, Rungta, Saladi, Schelten, Silva, Smith, Subramanian, Tan, Tang, Taylor, Williams, Kuan, Xu, Yan, Zarov, Zhang, Fan, Kambadur, Narang, Rodriguez, Stojnic, Edunov, and Scialom}]{touvron2023llama}
Hugo Touvron, Louis Martin, Kevin Stone, Peter Albert, Amjad Almahairi, Yasmine Babaei, Nikolay Bashlykov, Soumya Batra, Prajjwal Bhargava, Shruti Bhosale, Dan Bikel, Lukas Blecher, Cristian~Canton Ferrer, Moya Chen, Guillem Cucurull, David Esiobu, Jude Fernandes, Jeremy Fu, Wenyin Fu, Brian Fuller, Cynthia Gao, Vedanuj Goswami, Naman Goyal, Anthony Hartshorn, Saghar Hosseini, Rui Hou, Hakan Inan, Marcin Kardas, Viktor Kerkez, Madian Khabsa, Isabel Kloumann, Artem Korenev, Punit~Singh Koura, Marie-Anne Lachaux, Thibaut Lavril, Jenya Lee, Diana Liskovich, Yinghai Lu, Yuning Mao, Xavier Martinet, Todor Mihaylov, Pushkar Mishra, Igor Molybog, Yixin Nie, Andrew Poulton, Jeremy Reizenstein, Rashi Rungta, Kalyan Saladi, Alan Schelten, Ruan Silva, Eric~Michael Smith, Ranjan Subramanian, Xiaoqing~Ellen Tan, Binh Tang, Ross Taylor, Adina Williams, Jian~Xiang Kuan, Puxin Xu, Zheng Yan, Iliyan Zarov, Yuchen Zhang, Angela Fan, Melanie Kambadur, Sharan Narang, Aurelien Rodriguez, Robert Stojnic, Sergey Edunov, and Thomas
  Scialom. 2023.
\newblock \href {http://arxiv.org/abs/2307.09288} {Llama 2: Open foundation and fine-tuned chat models}.

\bibitem[{Whitehouse et~al.(2023)Whitehouse, Choudhury, and Aji}]{whitehouse-etal-2023-llm}
Chenxi Whitehouse, Monojit Choudhury, and Alham Aji. 2023.
\newblock \href {https://doi.org/10.18653/v1/2023.emnlp-main.44} {{LLM}-powered data augmentation for enhanced cross-lingual performance}.
\newblock In \emph{Proceedings of the 2023 Conference on Empirical Methods in Natural Language Processing}, pages 671--686, Singapore. Association for Computational Linguistics.

\bibitem[{Xue et~al.(2021)Xue, Constant, Roberts, Kale, Al-Rfou, Siddhant, Barua, and Raffel}]{xue-etal-2021-mt5}
Linting Xue, Noah Constant, Adam Roberts, Mihir Kale, Rami Al-Rfou, Aditya Siddhant, Aditya Barua, and Colin Raffel. 2021.
\newblock \href {https://doi.org/10.18653/v1/2021.naacl-main.41} {m{T}5: A massively multilingual pre-trained text-to-text transformer}.
\newblock In \emph{Proceedings of the 2021 Conference of the North American Chapter of the Association for Computational Linguistics: Human Language Technologies}, pages 483--498, Online. Association for Computational Linguistics.

\bibitem[{Üstün et~al.(2024)Üstün, Aryabumi, Yong, Ko, D'souza, Onilude, Bhandari, Singh, Ooi, Kayid, Vargus, Blunsom, Longpre, Muennighoff, Fadaee, Kreutzer, and Hooker}]{ustun2024aya}
Ahmet Üstün, Viraat Aryabumi, Zheng-Xin Yong, Wei-Yin Ko, Daniel D'souza, Gbemileke Onilude, Neel Bhandari, Shivalika Singh, Hui-Lee Ooi, Amr Kayid, Freddie Vargus, Phil Blunsom, Shayne Longpre, Niklas Muennighoff, Marzieh Fadaee, Julia Kreutzer, and Sara Hooker. 2024.
\newblock \href {http://arxiv.org/abs/2402.07827} {Aya model: An instruction finetuned open-access multilingual language model}.

\end{thebibliography}
